%% file: acl2023.tex
\pdfoutput=1

\documentclass[11pt]{article}

\usepackage[]{ACL2023}

\usepackage{times}
\usepackage{latexsym}

\usepackage[T1]{fontenc}

\usepackage[utf8]{inputenc}

\usepackage{microtype}

\usepackage{inconsolata}

\usepackage{graphicx}
\usepackage{multirow}
\usepackage{xcolor}
\usepackage{booktabs}
\usepackage{amssymb}
\usepackage{array}
\usepackage{amsmath}
\usepackage{stfloats}
\usepackage{adjustbox}
\usepackage{caption}
\usepackage{subcaption}
\usepackage{comment}
\usepackage{mathtools}

\newcommand\blfootnote[1]{%
  \begingroup
  \renewcommand\thefootnote{}\footnote{#1}%
  \addtocounter{footnote}{-1}%
  \endgroup
}

\title{KU-DMIS-MSRA at RadSum23: Pre-trained Vision-Language Model for Radiology Report Summarization}

\author{Gangwoo Kim$^{1 \diamond}$ \quad Hajung Kim$^{1}$ \quad Lei Ji$^{2}$ \quad Seongsu Bae$^{3}$ \quad Chanhwi Kim$^{1}$ \quad Mujeen Sung$^{1}$ \\
\quad \textbf{Hyunjae Kim}$^{1}$ \quad \textbf{Kun Yan}$^{4}$ \quad \textbf{Eric Chang}$^{5}$ \quad \textbf{Jaewoo Kang}$^{1,6\dagger}$ \\
$^{1}$Korea University$^{*}$
$^{2}$Microsoft Research Asia $^{3}$KAIST AI $^{4}$Beihang University \\
$^{5}$Centre for Perceptual and Interactive Intelligence, Hong Kong $^{6}$AIGEN Sciences Inc. \\
\texttt{\{gangwoo\_kim, kangj\}@korea.ac.kr}
}

\begin{document}
\maketitle

\blfootnote{\textsuperscript{$\diamond$} Work done while interning at Microsoft Research Asia }
\blfootnote{\textsuperscript{$\dagger$} Corresponding author }
\blfootnote{\textsuperscript{$*$} Department of Computer Science and Engineering }

\input{tabs/00_abstract}

\input{tabs/01_introduction}
\input{tabs/02_related_work}
\input{tabs/03_method}
\input{tabs/04_experiments}
\input{tabs/05_conclusion}

\section*{Acknowledgement}

This research was supported by the MSIT(Ministry of Science, ICT), Korea, under the High-Potential Individuals Global Training Program) (RS-2022-00155958) and the ICT Creative Consilience program(IITP-2023-2020-0-01819)  supervised by the IITP(Institute for Information \& Communications Technology Planning \& Evaluation).
It was also supported by the National Research Foundation of Korea, South Korea (NRF-2023R1A2C3004176) and a grant of the Korea Health Technology R\&D Project through the Korea Health Industry Development Institute (KHIDI), funded by the Ministry of Health \& Welfare, Republic of Korea (grant number: HR20C0021).

\bibliography{anthology,custom}
\bibliographystyle{acl_natbib}

\appendix



\end{document}

%% file: tabs/00_abstract.tex
\begin{abstract}
In this paper, we introduce CheXOFA, a new pre-trained vision-language model (VLM) for the chest X-ray domain. 
Our model is initially pre-trained on various multimodal datasets within the general domain before being transferred to the chest X-ray domain. 
Following a prominent VLM, we unify various domain-specific tasks into a simple sequence-to-sequence schema.
It enables the model to effectively learn the required knowledge and skills from limited resources in the domain.
Demonstrating superior performance on the benchmark datasets provided by the BioNLP shared task~\cite{DelbrouckRadSum23}, our model benefits from its training across multiple tasks and domains.
With subtle techniques including ensemble and factual calibration, our system achieves first place on the RadSum23 leaderboard for the hidden test set.

\end{abstract}

%% file: tabs/01_introduction.tex
\section{Introduction}

Chest radiography is a widely used imaging modality for assessing the thorax and diagnosing cardiopulmonary conditions. 
However, there is a significant shortage of clinical doctors in several under-resourced regions, delaying diagnosis and treatment and reducing the quality of care.
Developing an automated system for analyzing radiographs can improve radiologist workflow efficiency and expand healthcare services to these regions.

To promote research in this direction, the BioNLP 2023 workshop opens a new shared task called RadSum23~\cite{DelbrouckRadSum23} for the radiology report summarization.\footnote{\url{vilmedic.app/misc/bionlp23/sharedtask}}
In this task, participants are asked to build a model that takes a \textit{findings} section as input, which is a generic radiology report of a given X-ray image, and then outputs an \textit{impression} section, which is a summary of  key observations in the given report.
High-resolution X-ray images can be used as input along with findings sections.
The radiology report summarization task aims to effectively distill complex clinical observations from chest X-ray images into concise and coherent summaries. 


In this paper, we introduce CheXOFA (One For All tasks with Chest X-ray), a novel vision-language model designed for the chest X-ray domain. 
Specifically, we initialize our model using OFA~\cite{wang2022ofa}, a Transformer model pre-trained with a unified sequence-to-sequence schema on diverse uni- or cross-modal tasks such as image classification, language modeling, and image captioning.
We further train the model to generate full-text reports from the chest X-ray image using the MIMIC-CXR dataset~\cite{johnson2019mimic}.
Then, we fine-tune the model on the radiology report summarization task.
When summarizing the report, our model jointly encodes visual information from the chest X-ray image with the full-text report, taking advantage of its multimodal nature.
Additionally, we employ subtle techniques such as task-specific ensemble~\cite{dai-etal-2021-bdkg} and factual calibration to further improve the model performance. 


Our experiments demonstrate that our proposed methods largely enhance model performances on two test sets of the shared task.
On the official leaderboard \footnote{\url{vilmedic.app/misc/bionlp23/leaderboard/}} for MIMIC-CXR hidden test set, our system ranked first place, achieving the state-of-the-art performances in most evaluation metrics.
Especially, it surpasses the second-best model by 2.3 and 2.9 in BLEU and F1-CheXbert score.
In the ablation study, we demonstrate how much each method contributes to the improvement.




%% file: tabs/02_related_work.tex
\section{Related Works}
\subsection{Automated Radiology Report Generation}
Utilizing radiographic datasets such as MIMIC-CXR~\cite{johnson2019mimic} and CheXpert~\cite{irvin2019chexpert}, various methodologies for automated report generation have recently attracted attention. 
Notably, these datasets often include both chest X-ray images and free-text reports, enabling the use of automated rule-based labelers~\cite{irvin2019chexpert} or neural models~\cite{smit2020combining} to extract disease labels from the reports. 
It can be categorized into two tasks: 1) radiology report generation, which is similar to medical image captioning and aims to describe radiology images in detail (\textit{findings} section), having seen significant progress in recent years~\cite{chen2020generating, zhang2020radiology, liu2021contrastive, liu2021auto, miura2021improving, chen2022multi}; 
 2) radiology report summarization~\cite{zhang2018learning}, which focuses on summarizing \textit{findings} section into \textit{impressions} section in radiology reports. 
Most existing research~\cite{zhang2020optimizing, hu2021word, hu2022graph, karn2022differentiable} tends to focus on text-based summarization while some image-incorporating studies~\cite{delbrouck2021qiai, hu2022improving} use suboptimal methods lacking appropriate multi-modal pre-training objectives for the generative task.

\subsection{Multimodal Foundation Models}
Vision-language pretraining models are becoming the foundation models effective for multimodal tasks including Vilbert~\cite{lu2019vilbert},  OFA~\cite{wang2022ofa}, Flamingo~\cite{alayrac2022flamingo} for open-domain, and MedViLL~\cite{moon2022multi}, Clinical-BERT~\cite{yan2022clinical}, BioViL~\cite{10.1007/978-3-031-20059-5_1} and M3AE~\cite{chen2022multi} for biomedical domain. However, most of the existing works for clinical domain mainly focus on pretraining the encoder for understanding tasks like medical VQA, classification and so on, and rarely take this radiology report summarization task as downstream task. 
In this paper, we propose CheXOFA, a pre-trained generative VLM that learns the required capabilities for the radiology report summarization task. 

%% file: tabs/03_method.tex
\input{figs/overview}
\section{CheXOFA}
We propose a novel vision-language model, CheXOFA, specifically designed for the radiology report summarization.
We first initialize our model parameters with OFA~\cite{wang2022ofa}, which has been shown to be effective in the general domain. 
Then, we pre-train and fine-tune the model with the various tasks in the medical imaging domain. 
In addition, we newly introduce a factual calibration technique to further improve the model performance.


\subsection{Multimodal Architecture}
The backbone of the CheXOFA model is Transformer~\cite{vaswani2017attention} architecture with a sequence-to-sequence framework specialized for generative tasks.
Following BART~\cite{lewis2020bart} and GPT~\cite{radfordimproving}, we utilize byte-pair encoding (BPE)~\cite{sennrich2016neural} to transform text sequences into linguistic features of subword sequences.
On the vision side, we use a visual extractor to encode an image into a sequence of hidden representations.
Specifically, we divide an input image into fixed size of patches. Then, ResNet~\cite{he2016deep} modules are used to convolve the visual information into visual features $x^v \in \mathcal{R}^{|P| \times d}$, where $|P|$ is the number of patches and $d$ is the dimension of hidden representation. 
Overall, the linguistic features and the visual features are concatenated into one sequence for  feeding into the encoder-decoder Transformer for modality fusion and sequence generation. 

\subsection{Training and Inference}
Our model is optimized with the cross-entropy loss to the ground-truth sequences.
Given an input $x$ and an output $y$, we train model parameters $\theta$ by minimizing $\mathcal{L} = - \sum_{t=1}^{|y|} \log P_{\theta}(y_{t} | y_{<t}, x)$, where $x$ can be composed of instruction $x^i$, linguistic and visual features $x^l, x^v$.
In the inference phase, we choose the beam search as the decoding strategy to obtain better text sequences.
Additionally, we adjust the length penalty that assigns weights according to the length of each beam.
We control the output sequence length and explore the optimal value for it.

\subsection{Pre-training and Downstream Tasks}
Following \citet{wang2022ofa}, we unify cross-modal tasks into a simple sequence-to-sequence format.
We design a unified learning framework for pre-training and downstream tasks, which require multimodal reasoning ability.
CheXOFA's versatile design allows it to tackle a wide range of tasks using a single model. 
Furthermore, the model is designed with multitasking capabilities, allowing it to simultaneously handle multiple tasks across various modalities.
To this end, the model shares its parameters and schema across all tasks.
Meanwhile, we employ task-specific instructions manually crafted for each task.

We pre-train CheXOFA with the report generation (RGen) task and then fine-tune it with report summarization (RSum) on MIMIC-CXR dataset~\cite{johnson2019mimic}.
In the RGen task, the model learns to generate \textit{findings} section of the report, based on the chest X-ray image.
We use the same instruction $x^i$ with that of the image captioning task, ``What does the image describe?''.
For our target task, RSum, the model is trained to generate \textit{impression} section, given \textit{findings} section of the report.
We also exploit the corresponding chest X-ray image to jointly leverage the visual information.
Hence, an input $x$ is composed of visual features, subword tokens of \textit{findings} section, and instruction, ``what is the summary of the following article?''.
Furthermore, we newly design the classification-supported RSum task (cls-RSum) to enhance the factual correctness of the summary.
In the task, the model additionally performs a classification task for observed disease from the X-ray image or report. 
Then, it generates summaries based on the identified category, ensuring relevance and coherence.

\subsection{Ensemble with Factual Calibration}
To improve the overall performance, we utilize an ensemble method that combines various predictions from multiple models.
Following~\citet{dai-etal-2021-bdkg}, we select the best prediction based on the mutual similarity score.
In particular, we calculate similarity scores for every possible pair of predictions, creating a mutual similarity matrix. 
Subsequently, we aggregate the matrix in a row-wise fashion, averaging value for each row. 
The prediction with the highest score is selected as the final output. 
If multiple outputs have the same highest score, we randomly select the final output, ensuring that it is chosen in an unbiased manner. 
To compute the similarity, we use F1-RadGraph as a scoring function. 
Through the combination of diverse predictions, we are able to obtain an optimal summary and mitigate the failure of individual models. 

We perform a calibration to improve the factual correctness of the ensembled prediction.
We first extract medical observations from the prediction by using cheXbert labeler~\cite{smit2020combining}.
Then, we check whether they are matched with identified labels by the cls-RSum model. 
If not, the cls-RSum result is chosen as the final prediction.
We find the factual calibration performs effectively when the ensemble process yields a low aggregated similarity score, which is in proportion to the model's confidence.

%% file: figs/overview.tex
\begin{figure*} [t!]
    \centering
    \includegraphics[width=\textwidth]{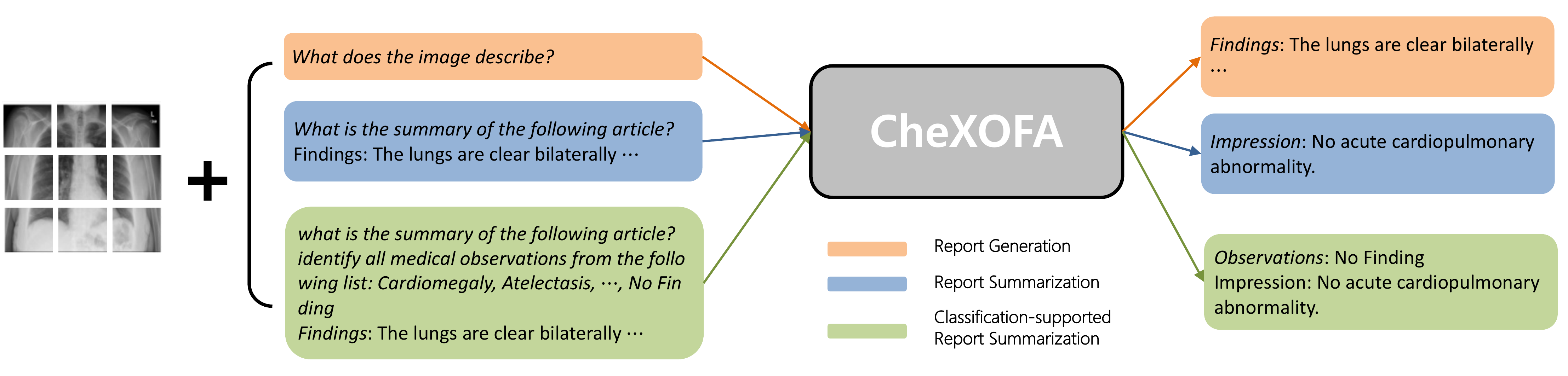}
    \caption{Overview of CheXOFA. It is pre-trained in the report generation task. We then fine-tune the pre-trained model in the report summarization task. We use task-specific instructions.
       }
    \label{fig:overview}
\end{figure*}

%% file: tabs/04_experiments.tex
\section{Experiment}
\input{tables/leaderboard_hidden}

\subsection{Experimental Setup}

\paragraph{Evaluation Metric}
For evaluation, we used five metrics: BLEU4~\citep{papineni-etal-2002-bleu}, ROUGE-L~\citep{lin-2004-rouge}, Bertscore~\citep{bert-score}, F1-cheXbert~\citep{zhang-etal-2020-optimizing}, and F1-RadGraph~\citep{delbrouck-etal-2022-improving}.
BLUE4 and ROUGE-L measure syntactical similarity based on n-gram overlap between reference and generated summaries, while Bertscore measures semantic similarity to handle synonyms and paraphrasing. 
To evaluate the factual correctness of generated summaries, F1-cheXbert and F1-RadGraph are used. 
Out of these five metrics, F1-RadGraph is selected as the primary metric for ranking participating systems, as it uses RadGraph~\citep{Jain2021RadGraphEC} annotations to better consider the consistency and correctness of extracted entities and relations.

\paragraph{Benchmark Datasets}
Our model was trained on MIMIC-CXR~\citep{johnson2019mimic} and evaluated on both the test set of MIMIC-CXR and a newly-collected hidden test set. 
MIMIC-CXR is a publicly available large dataset consisting of 128,032 report-image pairs with 227,835 multi-view images\footnote{The mentioned statistics are derived from reports containing both the \textit{findings} and \textit{impression} sections simultaneously.}. 
The dataset is split into training, validation, and test sets, which comprise 125,417, 991, and 1,624 image-report pairs, respectively. 
The MIMIC-CXR hidden test set was newly introduced in the RadSum23 challenge served by vilmedic~\cite{delbrouck-etal-2022-vilmedic}, and it comprises 1,000 out-of-domain image-report pairs collected from CheXpert images~\cite{irvin2019chexpert}\footnote{Although it comes from CheXpert dataset, we name it as MIMIC-CXR hidden test set, following the challenge description}.

\subsection{Results}
Table~\ref{table:leaderboard} shows the official results of the leaderboard on the hidden test set. 
Our model ranked first place among other systems on the leaderboard and achieved the state-of-the-art performance in four out of five evaluation metrics.
Especially, our model significantly outperformed the second-best model by 2.29 BLEU4 and 2.95 F1-cheXbert.

Table~\ref{table:leaderboard} also presents the official results on the public test set of MIMIC-CXR.
Our model achieved competitive performance with the best model in four out of five metrics and the best performance based on Bertscore. 
Overall results indicate that our method could generalize to diverse datasets, achieving outstanding performances on both hidden and visible test sets.
Conclusively, our method has remarkable capabilities to summarize radiology reports by capturing essential medical observations.


\input{tables/ablation}

\subsection{Ablation Study}
We performed an ablation study to analyze how each method contributes to the overall performance.
Table~\ref{table:ablation} shows the evaluation scores when removing each method from our best model on the test set of MIMIC-CXR.
Factual calibration improves the factual correctness scores, 0.3 of the F1-CheXbert score and 0.8 of F1-Radgraph score.
Using a single CheXOFA model shows a performance drop compared to the ensemble model by approximately 1.8 in F1-Radgraph.
Nevertheless, it achieves competitive performances with other participating systems.
Allowing the model to focus on the report (Text-only) achieves similar performance in ROUGE-L and F1-Radgraph scores relevant to the lexical overlap.
However, F1-cheXbert score significantly degrades, which indicates that models benefit from using multimodal information.
Fine-tuning a vanilla OFA model performs poorly in most scores, which shows the importance of the pre-training task.

%% file: tables/leaderboard_hidden.tex
\begin{table*}[t]

\footnotesize
\centering
\resizebox{0.95\textwidth}{!}{
\begin{tabular}{c|c|l|ccccc}
\toprule
\multirow{2}{*}{Track}  & \multirow{2}{*}{Rank} & \multirow{2}{*}{Team Name} & \multirow{2}{*}{BLEU} & \multirow{2}{*}{ROUGE} & \multirow{2}{*}{Bertscore} & CheXbert & Radgraph \\
 & & & & &  & F1 & F1 \\

\midrule \midrule
\multirow{5}{*}{Hidden Test}
& 1 & \textbf{ku-dmis-msra (ours)} & \textbf{18.62} & \underline{34.57} & \textbf{55.90} & \textbf{72.36} & \textbf{43.20} \\
& 2 & utsa-nlp & \underline{16.33} & \textbf{34.97} & \underline{55.54} & \underline{69.41} & \underline{42.86} \\
& 3 & knowlab & 14.41 & 33.63 & 54.72 & 67.20 & 39.98 \\
& 4 & shs-te-dti-mai & 14.59 & 32.43 & 53.99 & 6 8.99 & 38.40  \\
& 5 & aimi & 5.15 & 31.84 & 47.83 & 64.18 & 32.05  \\

\midrule
\multirow{5}{*}{Public Test}

& 1 & utsa-nlp & \textbf{25.87} & \textbf{47.86} & \underline{64.74} & \textbf{77.93} & \textbf{51.84}  \\
& 2 & \textbf{ku-dmis-msra (ours)} & \underline{25.58} & \underline{47.75} & \textbf{64.80} & \underline{76.29} & \underline{50.96}  \\
& 3 & knowlab  & 22.97 & 46.15 & 63.43 & 75.14 & 48.04 \\
& 4 & e-health csiro & 17.97 & 44.14 & 61.47 & 71.67 & 44.95 \\
& 5 & iuteam1 & 10.10 & 40.44 & 56.44 & 58.01 & 39.48 \\

\bottomrule
\end{tabular}
}
\caption{Official results of the leaderboards on MIMIC-CXR hidden test set and MIMIC-CXR test set. The Models are ranked based on F1-Radgraph score. The best score is displayed in bold typeface and the score of the second best model is underlined.
}\vspace{-0.3cm}
\label{table:leaderboard}
\end{table*}

%% file: tables/ablation.tex
\begin{table}[t!]
\footnotesize
\resizebox{0.48\textwidth}{!}{
    \begin{tabular}{lccc}
    \toprule
    \multirow{2}{*}{Model} & \multirow{2}{*}{ROUGE} & CheXbert & Radgraph  \\
     &  & F1 & F1  \\
    \midrule 
    CheXOFA (Ensem.) & 47.75 & 76.29 & 50.96   \\
    \quad w/o. Fact Calib. & 47.04 & 75.98 & 50.14   \\
    CheXOFA (Single) & 46.03 & 75.67 & 48.16   \\
    \quad Text-only & 46.30 & 73.59 & 47.78   \\
    \quad w/o. pre-training & 43.79 & 73.67 & 45.73   \\
    \bottomrule
    \end{tabular}
}
\caption{Ablation study on the public test set. We provide the performance of ensembled (Ensem.) and single results. Every component such as factual calibration (Fact Calib.), encoding multimodal inputs (Text-only), and pre-training contributes to the improvement.
}\vspace{-0.3cm}
\label{table:ablation}
\end{table}

%% file: tabs/05_conclusion.tex
\section{Conclusion}
We proposed a pre-trained VLM, CheXOFA, for the chest X-ray domain.
We showed that pre-traing with the report generation task improves the downstream task, report summarization.
Taking advantage of its multimodal nature, we improved the performance by jointly encoding visual and linguistic features.
Furthermore, we explored subtle techniques such as ensemble and factual calibration to improve the model performance.
Our experimental results demonstrated that proposed methods benefit the summarization performance.
Our model ranked first on the hidden test set in RadSum23 shared task.
We showed promising results about the domain-specific VLM in the chest X-ray tasks. 
We hope that our method can shed light on automating radiology report generation.

%% file: acl2023.bbl
\begin{thebibliography}{36}
\expandafter\ifx\csname natexlab\endcsname\relax\def\natexlab#1{#1}\fi

\bibitem[{Alayrac et~al.(2022)Alayrac, Donahue, Luc, Miech, Barr, Hasson, Lenc,
  Mensch, Millican, Reynolds et~al.}]{alayrac2022flamingo}
Jean-Baptiste Alayrac, Jeff Donahue, Pauline Luc, Antoine Miech, Iain Barr,
  Yana Hasson, Karel Lenc, Arthur Mensch, Katherine Millican, Malcolm Reynolds,
  et~al. 2022.
\newblock Flamingo: a visual language model for few-shot learning.
\newblock \emph{Advances in Neural Information Processing Systems},
  35:23716--23736.

\bibitem[{Boecking et~al.(2022)Boecking, Usuyama, Bannur, Castro, Schwaighofer,
  Hyland, Wetscherek, Naumann, Nori, Alvarez-Valle, Poon, and
  Oktay}]{10.1007/978-3-031-20059-5_1}
Benedikt Boecking, Naoto Usuyama, Shruthi Bannur, Daniel~C. Castro, Anton
  Schwaighofer, Stephanie Hyland, Maria Wetscherek, Tristan Naumann, Aditya
  Nori, Javier Alvarez-Valle, Hoifung Poon, and Ozan Oktay. 2022.
\newblock Making the most of text semantics to improve biomedical
  vision--language processing.
\newblock In \emph{Computer Vision -- ECCV 2022}, pages 1--21, Cham. Springer
  Nature Switzerland.

\bibitem[{Chen et~al.(2022)Chen, Du, Hu, Liu, Li, Wan, and
  Chang}]{chen2022multi}
Zhihong Chen, Yuhao Du, Jinpeng Hu, Yang Liu, Guanbin Li, Xiang Wan, and
  Tsung-Hui Chang. 2022.
\newblock Multi-modal masked autoencoders for medical vision-and-language
  pre-training.
\newblock In \emph{Medical Image Computing and Computer Assisted
  Intervention--MICCAI 2022: 25th International Conference, Singapore,
  September 18--22, 2022, Proceedings, Part V}, pages 679--689. Springer.

\bibitem[{Chen et~al.(2020)Chen, Song, Chang, and Wan}]{chen2020generating}
Zhihong Chen, Yan Song, Tsung-Hui Chang, and Xiang Wan. 2020.
\newblock Generating radiology reports via memory-driven transformer.
\newblock In \emph{Proceedings of the 2020 Conference on Empirical Methods in
  Natural Language Processing (EMNLP)}, pages 1439--1449.

\bibitem[{Dai et~al.(2021)Dai, Wang, Lyu, and Zhu}]{dai-etal-2021-bdkg}
Songtai Dai, Quan Wang, Yajuan Lyu, and Yong Zhu. 2021.
\newblock \href {https://doi.org/10.18653/v1/2021.bionlp-1.11} {{BDKG} at
  {MEDIQA} 2021: System report for the radiology report summarization task}.
\newblock In \emph{Proceedings of the 20th Workshop on Biomedical Language
  Processing}, pages 103--111, Online. Association for Computational
  Linguistics.

\bibitem[{Delbrouck et~al.(2022{\natexlab{a}})Delbrouck, Chambon, Bluethgen,
  Tsai, Almusa, and Langlotz}]{delbrouck-etal-2022-improving}
Jean-Benoit Delbrouck, Pierre Chambon, Christian Bluethgen, Emily Tsai, Omar
  Almusa, and Curtis Langlotz. 2022{\natexlab{a}}.
\newblock \href {https://aclanthology.org/2022.findings-emnlp.319} {Improving
  the factual correctness of radiology report generation with semantic
  rewards}.
\newblock In \emph{Findings of the Association for Computational Linguistics:
  EMNLP 2022}, pages 4348--4360, Abu Dhabi, United Arab Emirates. Association
  for Computational Linguistics.

\bibitem[{Delbrouck et~al.(2022{\natexlab{b}})Delbrouck, Saab, Varma, Eyuboglu,
  Chambon, Dunnmon, Zambrano, Chaudhari, and
  Langlotz}]{delbrouck-etal-2022-vilmedic}
Jean-benoit Delbrouck, Khaled Saab, Maya Varma, Sabri Eyuboglu, Pierre Chambon,
  Jared Dunnmon, Juan Zambrano, Akshay Chaudhari, and Curtis Langlotz.
  2022{\natexlab{b}}.
\newblock \href {https://doi.org/10.18653/v1/2022.acl-demo.3} {{V}i{LM}edic: a
  framework for research at the intersection of vision and language in medical
  {AI}}.
\newblock In \emph{Proceedings of the 60th Annual Meeting of the Association
  for Computational Linguistics: System Demonstrations}, pages 23--34, Dublin,
  Ireland. Association for Computational Linguistics.

\bibitem[{Delbrouck et~al.(2023)Delbrouck, Varma, Chambon, and
  Langlotz}]{DelbrouckRadSum23}
Jean-Benoit Delbrouck, Maya Varma, Pierre Chambon, and Curtis Langlotz. 2023.
\newblock Overview of the radsum23 shared task on multi-modal and
  multi-anatomical radiology report summarization.
\newblock In \emph{Proceedings of the 22st Workshop on Biomedical Language
  Processing}, Toronto, Canada. Association for Computational Linguistics.

\bibitem[{Delbrouck et~al.(2021)Delbrouck, Zhang, and
  Rubin}]{delbrouck2021qiai}
Jean-Benoit Delbrouck, Cassie Zhang, and Daniel Rubin. 2021.
\newblock Qiai at mediqa 2021: Multimodal radiology report summarization.
\newblock In \emph{Proceedings of the 20th Workshop on Biomedical Language
  Processing}, pages 285--290.

\bibitem[{He et~al.(2016)He, Zhang, Ren, and Sun}]{he2016deep}
Kaiming He, Xiangyu Zhang, Shaoqing Ren, and Jian Sun. 2016.
\newblock Deep residual learning for image recognition.
\newblock In \emph{Proceedings of the IEEE conference on computer vision and
  pattern recognition}, pages 770--778.

\bibitem[{Hu et~al.(2022{\natexlab{a}})Hu, Chen, Liu, Wan, and
  Chang}]{hu2022improving}
Jinpeng Hu, Zhihong Chen, Yang Liu, Xiang Wan, and Tsung-Hui Chang.
  2022{\natexlab{a}}.
\newblock Improving radiology summarization with radiograph and anatomy
  prompts.
\newblock \emph{arXiv preprint arXiv:2210.08303}.

\bibitem[{Hu et~al.(2021)Hu, Li, Chen, Shen, Song, Wan, and Chang}]{hu2021word}
Jinpeng Hu, Jianling Li, Zhihong Chen, Yaling Shen, Yan Song, Xiang Wan, and
  Tsung-Hui Chang. 2021.
\newblock Word graph guided summarization for radiology findings.
\newblock In \emph{Findings of the Association for Computational Linguistics:
  ACL-IJCNLP 2021}, pages 4980--4990.

\bibitem[{Hu et~al.(2022{\natexlab{b}})Hu, Li, Chen, Li, Wan, and
  Chang}]{hu2022graph}
Jinpeng Hu, Zhuo Li, Zhihong Chen, Zhen Li, Xiang Wan, and Tsung-Hui Chang.
  2022{\natexlab{b}}.
\newblock Graph enhanced contrastive learning for radiology findings
  summarization.
\newblock In \emph{Proceedings of the 60th Annual Meeting of the Association
  for Computational Linguistics (Volume 1: Long Papers)}, pages 4677--4688.

\bibitem[{Irvin et~al.(2019)Irvin, Rajpurkar, Ko, Yu, Ciurea-Ilcus, Chute,
  Marklund, Haghgoo, Ball, Shpanskaya et~al.}]{irvin2019chexpert}
Jeremy Irvin, Pranav Rajpurkar, Michael Ko, Yifan Yu, Silviana Ciurea-Ilcus,
  Chris Chute, Henrik Marklund, Behzad Haghgoo, Robyn Ball, Katie Shpanskaya,
  et~al. 2019.
\newblock Chexpert: A large chest radiograph dataset with uncertainty labels
  and expert comparison.
\newblock In \emph{Proceedings of the AAAI conference on artificial
  intelligence}, volume~33, pages 590--597.

\bibitem[{Jain et~al.(2021)Jain, Agrawal, Saporta, Truong, Duong, Bui, Chambon,
  Zhang, Lungren, Ng, Langlotz, and Rajpurkar}]{Jain2021RadGraphEC}
Saahil Jain, Ashwin Agrawal, Adriel Saporta, Steven Truong, D.~Duong, Tan Bui,
  Pierre Chambon, Yuhao Zhang, Matthew~P. Lungren, Andrew~Y. Ng, Curt~P.
  Langlotz, and Pranav Rajpurkar. 2021.
\newblock Radgraph: Extracting clinical entities and relations from radiology
  reports.
\newblock \emph{ArXiv}, abs/2106.14463.

\bibitem[{Johnson et~al.(2019)Johnson, Pollard, Berkowitz, Greenbaum, Lungren,
  Deng, Mark, and Horng}]{johnson2019mimic}
Alistair~EW Johnson, Tom~J Pollard, Seth~J Berkowitz, Nathaniel~R Greenbaum,
  Matthew~P Lungren, Chih-ying Deng, Roger~G Mark, and Steven Horng. 2019.
\newblock Mimic-cxr, a de-identified publicly available database of chest
  radiographs with free-text reports.
\newblock \emph{Scientific data}, 6(1):317.

\bibitem[{Karn et~al.(2022)Karn, Liu, Sch{\"u}tze, and
  Farri}]{karn2022differentiable}
Sanjeev~Kumar Karn, Ning Liu, Hinrich Sch{\"u}tze, and Oladimeji Farri. 2022.
\newblock Differentiable multi-agent actor-critic for multi-step radiology
  report summarization.
\newblock In \emph{Proceedings of the 60th Annual Meeting of the Association
  for Computational Linguistics (Volume 1: Long Papers)}, pages 1542--1553.

\bibitem[{Lewis et~al.(2020)Lewis, Liu, Goyal, Ghazvininejad, Mohamed, Levy,
  Stoyanov, and Zettlemoyer}]{lewis2020bart}
Mike Lewis, Yinhan Liu, Naman Goyal, Marjan Ghazvininejad, Abdelrahman Mohamed,
  Omer Levy, Veselin Stoyanov, and Luke Zettlemoyer. 2020.
\newblock Bart: Denoising sequence-to-sequence pre-training for natural
  language generation, translation, and comprehension.
\newblock In \emph{Proceedings of the 58th Annual Meeting of the Association
  for Computational Linguistics}, pages 7871--7880.

\bibitem[{Lin(2004)}]{lin-2004-rouge}
Chin-Yew Lin. 2004.
\newblock \href {https://aclanthology.org/W04-1013} {{ROUGE}: A package for
  automatic evaluation of summaries}.
\newblock In \emph{Text Summarization Branches Out}, pages 74--81, Barcelona,
  Spain. Association for Computational Linguistics.

\bibitem[{Liu et~al.(2021{\natexlab{a}})Liu, Yin, Wu, Ge, Zhang, and
  Sun}]{liu2021contrastive}
Fenglin Liu, Changchang Yin, Xian Wu, Shen Ge, Ping Zhang, and Xu~Sun.
  2021{\natexlab{a}}.
\newblock Contrastive attention for automatic chest x-ray report generation.
\newblock In \emph{Findings of the Association for Computational Linguistics:
  ACL-IJCNLP 2021}, pages 269--280.

\bibitem[{Liu et~al.(2021{\natexlab{b}})Liu, You, Wu, Ge, Sun
  et~al.}]{liu2021auto}
Fenglin Liu, Chenyu You, Xian Wu, Shen Ge, Xu~Sun, et~al. 2021{\natexlab{b}}.
\newblock Auto-encoding knowledge graph for unsupervised medical report
  generation.
\newblock \emph{Advances in Neural Information Processing Systems},
  34:16266--16279.

\bibitem[{Lu et~al.(2019)Lu, Batra, Parikh, and Lee}]{lu2019vilbert}
Jiasen Lu, Dhruv Batra, Devi Parikh, and Stefan Lee. 2019.
\newblock Vilbert: Pretraining task-agnostic visiolinguistic representations
  for vision-and-language tasks.
\newblock \emph{Advances in neural information processing systems}, 32.

\bibitem[{Miura et~al.(2021)Miura, Zhang, Tsai, Langlotz, and
  Jurafsky}]{miura2021improving}
Yasuhide Miura, Yuhao Zhang, Emily Tsai, Curtis Langlotz, and Dan Jurafsky.
  2021.
\newblock Improving factual completeness and consistency of image-to-text
  radiology report generation.
\newblock In \emph{Proceedings of the 2021 Conference of the North American
  Chapter of the Association for Computational Linguistics: Human Language
  Technologies}, pages 5288--5304.

\bibitem[{Moon et~al.(2022)Moon, Lee, Shin, Kim, and Choi}]{moon2022multi}
Jong~Hak Moon, Hyungyung Lee, Woncheol Shin, Young-Hak Kim, and Edward Choi.
  2022.
\newblock Multi-modal understanding and generation for medical images and text
  via vision-language pre-training.
\newblock \emph{IEEE Journal of Biomedical and Health Informatics},
  26(12):6070--6080.

\bibitem[{Papineni et~al.(2002)Papineni, Roukos, Ward, and
  Zhu}]{papineni-etal-2002-bleu}
Kishore Papineni, Salim Roukos, Todd Ward, and Wei-Jing Zhu. 2002.
\newblock \href {https://doi.org/10.3115/1073083.1073135} {{B}leu: a method for
  automatic evaluation of machine translation}.
\newblock In \emph{Proceedings of the 40th Annual Meeting of the Association
  for Computational Linguistics}, pages 311--318, Philadelphia, Pennsylvania,
  USA. Association for Computational Linguistics.

\bibitem[{Radford et~al.()Radford, Narasimhan, Salimans, and
  Sutskever}]{radfordimproving}
Alec Radford, Karthik Narasimhan, Tim Salimans, and Ilya Sutskever.
\newblock Improving language understanding by generative pre-training.

\bibitem[{Sennrich et~al.(2016)Sennrich, Haddow, and
  Birch}]{sennrich2016neural}
Rico Sennrich, Barry Haddow, and Alexandra Birch. 2016.
\newblock Neural machine translation of rare words with subword units.
\newblock In \emph{Proceedings of the 54th Annual Meeting of the Association
  for Computational Linguistics (Volume 1: Long Papers)}, pages 1715--1725.

\bibitem[{Smit et~al.(2020)Smit, Jain, Rajpurkar, Pareek, Ng, and
  Lungren}]{smit2020combining}
Akshay Smit, Saahil Jain, Pranav Rajpurkar, Anuj Pareek, Andrew~Y Ng, and
  Matthew Lungren. 2020.
\newblock Combining automatic labelers and expert annotations for accurate
  radiology report labeling using bert.
\newblock In \emph{Proceedings of the 2020 Conference on Empirical Methods in
  Natural Language Processing (EMNLP)}, pages 1500--1519.

\bibitem[{Vaswani et~al.(2017)Vaswani, Shazeer, Parmar, Uszkoreit, Jones,
  Gomez, Kaiser, and Polosukhin}]{vaswani2017attention}
Ashish Vaswani, Noam Shazeer, Niki Parmar, Jakob Uszkoreit, Llion Jones,
  Aidan~N Gomez, {\L}ukasz Kaiser, and Illia Polosukhin. 2017.
\newblock Attention is all you need.
\newblock \emph{Advances in neural information processing systems}, 30.

\bibitem[{Wang et~al.(2022)Wang, Yang, Men, Lin, Bai, Li, Ma, Zhou, Zhou, and
  Yang}]{wang2022ofa}
Peng Wang, An~Yang, Rui Men, Junyang Lin, Shuai Bai, Zhikang Li, Jianxin Ma,
  Chang Zhou, Jingren Zhou, and Hongxia Yang. 2022.
\newblock Ofa: Unifying architectures, tasks, and modalities through a simple
  sequence-to-sequence learning framework.
\newblock In \emph{International Conference on Machine Learning}, pages
  23318--23340. PMLR.

\bibitem[{Yan and Pei(2022)}]{yan2022clinical}
Bin Yan and Mingtao Pei. 2022.
\newblock Clinical-bert: Vision-language pre-training for radiograph diagnosis
  and reports generation.
\newblock In \emph{Proceedings of the AAAI Conference on Artificial
  Intelligence}, volume~36, pages 2982--2990.

\bibitem[{Zhang et~al.(2020{\natexlab{a}})Zhang, Kishore, Wu, Weinberger, and
  Artzi}]{bert-score}
Tianyi Zhang, Varsha Kishore, Felix Wu, Kilian~Q. Weinberger, and Yoav Artzi.
  2020{\natexlab{a}}.
\newblock \href {https://openreview.net/forum?id=SkeHuCVFDr} {Bertscore:
  Evaluating text generation with bert}.
\newblock In \emph{International Conference on Learning Representations}.

\bibitem[{Zhang et~al.(2020{\natexlab{b}})Zhang, Wang, Xu, Yu, Yuille, and
  Xu}]{zhang2020radiology}
Yixiao Zhang, Xiaosong Wang, Ziyue Xu, Qihang Yu, Alan Yuille, and Daguang Xu.
  2020{\natexlab{b}}.
\newblock When radiology report generation meets knowledge graph.
\newblock In \emph{Proceedings of the AAAI Conference on Artificial
  Intelligence}, volume~34, pages 12910--12917.

\bibitem[{Zhang et~al.(2018)Zhang, Ding, Qian, Manning, and
  Langlotz}]{zhang2018learning}
Yuhao Zhang, Daisy~Yi Ding, Tianpei Qian, Christopher~D Manning, and Curtis~P
  Langlotz. 2018.
\newblock Learning to summarize radiology findings.
\newblock In \emph{Proceedings of the Ninth International Workshop on Health
  Text Mining and Information Analysis}, pages 204--213.

\bibitem[{Zhang et~al.(2020{\natexlab{c}})Zhang, Merck, Tsai, Manning, and
  Langlotz}]{zhang2020optimizing}
Yuhao Zhang, Derek Merck, Emily Tsai, Christopher~D Manning, and Curtis
  Langlotz. 2020{\natexlab{c}}.
\newblock Optimizing the factual correctness of a summary: A study of
  summarizing radiology reports.
\newblock In \emph{Proceedings of the 58th Annual Meeting of the Association
  for Computational Linguistics}, pages 5108--5120.

\bibitem[{Zhang et~al.(2020{\natexlab{d}})Zhang, Merck, Tsai, Manning, and
  Langlotz}]{zhang-etal-2020-optimizing}
Yuhao Zhang, Derek Merck, Emily Tsai, Christopher~D. Manning, and Curtis
  Langlotz. 2020{\natexlab{d}}.
\newblock \href {https://doi.org/10.18653/v1/2020.acl-main.458} {Optimizing the
  factual correctness of a summary: A study of summarizing radiology reports}.
\newblock In \emph{Proceedings of the 58th Annual Meeting of the Association
  for Computational Linguistics}, pages 5108--5120, Online. Association for
  Computational Linguistics.

\end{thebibliography}
